\documentclass[conference]{IEEEtran}
\IEEEoverridecommandlockouts
\usepackage{cite}
\usepackage{amsmath,amssymb,amsfonts}
\usepackage{algorithmic}
\usepackage{graphicx}
\usepackage{balance}
\usepackage{textcomp}
\usepackage{xcolor}
\usepackage{hyperref}
\hypersetup{
    colorlinks=true,
    linkcolor=orange,    
    urlcolor=orange,      
    citecolor=orange,
    pdfusetitle=true
}

\usepackage[sort&compress,numbers]{natbib}
\def\BibTeX{{\rm B\kern-.05em{\sc i\kern-.025em b}\kern-.08em
    T\kern-.1667em\lower.7ex\hbox{E}\kern-.125emX}}

\newcommand{\linebreakand}{%
  \end{@IEEEauthorhalign}
  \hfill\mbox{}\par
  \mbox{}\hfill\begin{@IEEEauthorhalign}
}

\begin{document}

\title{Shared Autonomy for Proximal Teaching\\

}

\author{
\IEEEauthorblockN{Megha Srivastava*\thanks{*Both authors contributed equally. We make source code available at \texttt{\href{https://github.com/Stanford-ILIAD/proximal-teaching}{https://github.com/Stanford-ILIAD/proximal-teaching}.}}}
\IEEEauthorblockA{\textit{Computer Science Department} \\
\textit{Stanford University}\\
Stanford, USA\\
megha@cs.stanford.edu}
\and
\IEEEauthorblockN{Reihaneh Iranmanesh*}
\IEEEauthorblockA{\textit{Computer Science Department} \\
\textit{Amherst College}\\
Amherst, USA\\
riranmanesh25@amherst.edu}
\and
\IEEEauthorblockN{Yuchen Cui}
\IEEEauthorblockA{\textit{Computer Science Department} \\
\textit{University of California Los Angeles}\\
Los Angeles, USA \\
yuchencui@cs.ucla.edu}
\and
\IEEEauthorblockN{Deepak Gopinath}
\IEEEauthorblockA{\textit{Human Interactive Driving} \\
\textit{Toyota Research Institute}\\
Cambridge, USA \\
deepak.gopinath@tri.global}
\and
\IEEEauthorblockN{Emily Sarah Sumner}
\IEEEauthorblockA{\textit{Human Interactive Driving} \\
\textit{Toyota Research Institute}\\
Cambridge, USA \\
emily.sumner@tri.global}
\and
\IEEEauthorblockN{Andrew Silva}
\IEEEauthorblockA{\textit{Human Interactive Driving} \\
\textit{Toyota Research Institute}\\
Cambridge, USA \\
andrew.silva@gatech.edu}
\and
\IEEEauthorblockN{Laporsha Dees}
\IEEEauthorblockA{\textit{Human Interactive Driving} \\
\textit{Toyota Research Institute}\\
Cambridge, USA \\
laporsha.dees.ctr@tri.global}
\and
\IEEEauthorblockN{\textcolor{white}{Guy Rosman}}
\IEEEauthorblockA{\textit{\textcolor{white}{Human Interactive Driving}} \\
\textit{\textcolor{white}{Toyota Research Institute}}\\
\textcolor{white}{Cambridge, USA} \\
\textcolor{white}{guy.rosman@tri.global}}
\and
\IEEEauthorblockN{Guy Rosman}
\IEEEauthorblockA{\textit{Human Interactive Driving} \\
\textit{Toyota Research Institute}\\
Cambridge, USA \\
guy.rosman@tri.global}
\and
\IEEEauthorblockN{Dorsa Sadigh}
\IEEEauthorblockA{\textit{Computer Science Department} \\
\textit{Stanford University}\\
Stanford, USA \\
dorsa@cs.stanford.edu}
\and
\IEEEauthorblockN{\textcolor{white}{Guy Rosman}}
\IEEEauthorblockA{\textit{\textcolor{white}{Human Interactive Driving}} \\
\textit{\textcolor{white}{Toyota Research Institute}}\\
\textcolor{white}{Cambridge, USA} \\
\textcolor{white}{guy.rosman@tri.global}}
}

\maketitle
\begin{abstract}
Motor skill learning often requires experienced professionals who can provide personalized instruction. Unfortunately, the availability of high-quality training can be limited for specialized tasks, such as high performance racing. Several recent works have leveraged AI-assistance to improve instruction of tasks ranging from rehabilitation to surgical robot tele-operation. However, these works often make simplifying assumptions on the student learning process, and fail to model how a teacher's assistance interacts with different individuals' abilities when determining optimal teaching strategies. Inspired by the idea of scaffolding from educational psychology, we leverage shared autonomy, a framework for combining user inputs with robot autonomy, to aid with curriculum design. Our key insight is that the way a student's behavior improves in the presence of assistance from an autonomous agent can highlight which sub-skills might be most ``learnable'' for the student, or within their Zone of Proximal Development. We use this to design \texttt{Z-COACH}, a method for using shared autonomy to provide personalized instruction targeting interpretable task sub-skills. In a user study (n=50), where we teach high performance racing in a simulated environment of the Thunderhill Raceway Park with the CARLA Autonomous Driving simulator, we show that \texttt{Z-COACH} helps identify which skills each student should first practice, leading to an overall improvement in driving time, behavior, and smoothness. Our work shows that increasingly available semi-autonomous capabilities (e.g. in vehicles, robots) can not only assist human users, but also help \textit{teach} them. 
\end{abstract}


\section{Introduction}
\begin{figure}
\centerline{\includegraphics[width=0.375\textwidth]{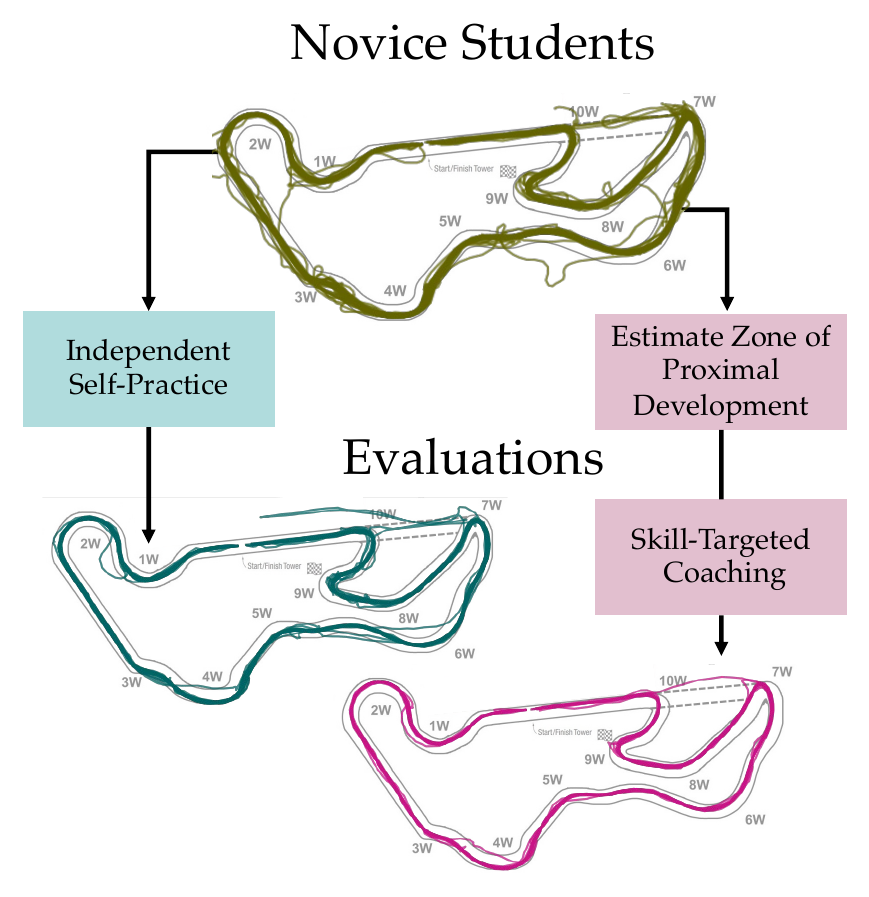}}
\caption{We evaluate \texttt{Z-COACH} on a high performance racing task in a simulated environment of the Thunderhill Raceway Park. \texttt{Z-COACH} identifies which task skills an individual student can only perform with assistance (i.e. within their Zone of Proximal Development), and then provides targeted coaching via skill-focused shared autonomy. Students receiving coaching from \texttt{Z-COACH} generally learned smoother racing lines than students practicing independently for the equivalent amount of time, as shown above by overlaying the trajectories from all participants in our human subject study ($n=50$).  }
\label{fig:pull}
\end{figure}


Imagine a young adult who wishes to learn how to drive, borrowing her family's semi-autonomous vehicle to practice in their neighborhood. In real time, the vehicle estimates the skill capabilities and quality of this student's driving, while also identifying what driving skills are required for the road ahead. As she encounters novel driving situations, the vehicle's semi-autonomous control features are adjusted to ensure the driver is appropriately challenged. This assistance is personalized; while she may be overly cautious and needs to learn to let go of the brake pedal, another novice driver might be overconfident and stay constantly at a high throttle. Over time, as the driver achieves proficiency in one skill (e.g. steering),  the vehicle is able to identify and assist her in learning more complex control maneuvers. In this shared autonomy setting, the vehicle is providing tailored assistance not only for safety and comfort, but also to make sure the student \textit{improves} their driving skills. 


Shared autonomous control is a promising paradigm for humans to achieve near-optimal task performance by offloading some decision-making to an autonomous agent \citep{reddy2018shared, aigner1997shared}. However, similar to how generative AI tools can have negative long-term effects on student learning \cite{bastani2024generative}, a significant concern with shared autonomy is the gradual loss of human control skills due to over-reliance on intelligent and assistive systems~\cite{de-Winter2023-cp}. While prior work has proposed ``learning-compatible'' forms of shared autonomy \cite{bragg2020fake}, they do not consider complex domains where learning entails a structured progression through a hierarchy of skills (e.g. learning steering before mastering sharp turns).

Inspired by the notion of scaffolding within the broader education literature, we propose optimizing the form of intelligent assistance to enhance skill development at a level that is appropriately challenging for a student, aligned with the concept of the Zone of Proximal Development (ZPD)~\cite{vygotsky1978mind}. The ZPD is loosely defined as the gap between a student's actual developmental level, determined from independent problem-solving, and the level of potential development, based on their performance when problem-solving	with assistance ~\cite{vygotsky1978mind}. However, the impact of the \textit{type} of assistance within scaffolding teaching has traditionally been overlooked, often taking the form of simple interventions such as verbal feedback \cite{luckin1999ecolab}. In light of the increasing interaction between autonomous systems and humans, we take a new perspective with shared autonomy: can the way a student's behavior changes with AI assistance help inform an appropriate learning curriculum?

We propose \texttt{Z-COACH}, a framework for leveraging shared autonomy to aid with both student modeling (i.e. identifying what skills are within a student's ZPD) and coaching (i.e. helping a student improve at a skill) for arbitrary motor control tasks. Unlike prior work on AI-assisted coaching that guides students to practice the skill they find most difficult \cite{srivastava2022motor}, \texttt{Z-COACH} uses shared autonomy to identify skills which a student is most likely to improve at a given point in time. We then apply \texttt{Z-COACH} to the task of high performance racing (HPR) in simulation, and conduct a user study $(n=50)$ demonstrating that \texttt{Z-COACH} helps improve a student's driving time, behavior, and smoothness in comparison to a self-practice baseline. 
Our main contributions include:
\begin{itemize}
\item An approach to characterize a student's ZPD based on a shared autonomy assistance for motor control tasks, and a formulation of ZPD that take both assisted and unassisted student performance into account.
\item A method to decide when to apply shared autonomy to help a student improve a particular skill, based on improving the interpretability of existing unsupervised skill discovery algorithms used in \cite{srivastava2022motor}.
\item Demonstration of the proposed framework in a human-in-the-loop experiment with the CARLA Autonomous Driving simulator, demonstrating how optimizing the shared autonomy assistance based on the student's estimated ZPD results in greater student improvement in a high performance race training session based on the Thunderhill Raceway Park.
\end{itemize}

\section{Related Work}
\subsection{Shared Autonomy}

Existing research on shared autonomy includes developing adaptive control algorithms that adapt autonomy levels based on context or user performance~\cite{abbink2018topology,xing2020driver,gopinath2020active,zurek2021situational}, designing intuitive human-machine interfaces~\cite{da2021biasing,chen2022mirror}, and studying the impact of shared control on learning and skill acquisition~\cite{yu2023coach}. 
Applications span rehabilitation robotics assisting patients in regaining motor functions~\cite{saadah2004autonomy,okamura2010medical,argall2018autonomy}, 
remote teleoperation systems ~\cite{mower2021sharedcontrol}, 
advanced driving systems where control is shared between the driver and autonomous features to improve safety~\cite{wang2020review, marcano2020review, reitmann2024shared},
assistive devices for individuals with disabilities~\cite{losey2020latent,karamcheti2022lila,cui2023no,udupa2023shared}, 
and flight training simulators~\cite{byeon2024flight}. 
Another line of research explores machine learning techniques to enhance the efficiency and safety of shared control systems~\cite{li2019shared}. We refer to  \cite{wang2020review} for a survey of the full literature, which is beyond the scope of this paper.





Within the human-robot interaction community, researchers have approached shared autonomy from a reinforcement learning perspective \cite{reddy2018shared}, sought to  perform inference over human preference and intent \cite{nemlekar2023robot, bobu2020less} as well as explicitly reason about human performance limits and the diverse state distributions that would be conducive for learning \cite{bragg2020fake}. However, these works do not explicitly account for scaffolding in learning. 

Closest to our work is the work of \citet{byeon2024flight}, which uses the Mahalanobis distance between novice and expert trajectories to update the parameters of a shared autonomy-based control for training novices in an urban air mobility task. In contrast, we propose leveraging shared autonomy not just during coaching, but also as a way to estimate which sub-skills of a task are most conducive to learning.

\subsection{Student Modeling and Pedagogy}
Research on student modeling focuses on how to effectively model a student's learning capabilities in order to tailor personalized learning experiences.
This includes methods like Bayesian Knowledge Tracing to predict student knowledge states and performance~\cite{corbett2005kt, david2016sequencing}, as well as models of pedagogy, such as how demonstrators choose to act when teaching a student new skills \citep{ho2016showing, ho2018effectively, csibra2009pedagogy}. Two pedagogical concepts relevant to our work are \textit{interleaving}, where students practice multiple skills at once, and \textit{scaffolding}, where teachers provide temporary support to help students develop new skills \cite{vygotsky1978mind}.

\paragraph{Teaching Human Students}
There exists a large amount of interest in developing educational robot technology to aid with human learning. For example, \citet{chen2024integrating} studies how social robots can play a role in teaching scenarios as coaches or mock-students, and designers of Doodlebot, a mobile social robot, sought to provide scaffolding that encouraged students to draw more creatively \citep{william2024doodlebot}. A seperate line of work seeks to model human drivers' systematic suboptimality by comparing students to an optimal trajectory from an MPC controller to identify if students are under-steering, under-speed, over-steering, or over-speed~\cite{schrum2022mind,schrum2022reciprocal}. Finally, recent research  explores how robots can effectively collaborate with humans in cooperative tasks by adopting roles that facilitate human learning ~\cite{hou2023teachingbot,yu2023coach}. In contrast to these works, we seek to use shared autonomy to more explicitly consider a student's ZPD in complex tasks that require learning multiple skills.

\paragraph{Modeling the Zone of Proximal Development}
Several non-robotic systems explicitly leverage Vygotsky's Zone of Proximal Development (ZPD)—the gap between what a learner can do independently and with assistance~\cite{vygotsky1978mind}.  These include intelligent tutoring systems that use these models to adjust instruction difficulty within the learner's ZPD, providing appropriate challenges and support~\cite{clement2015multi, vainas2019gotsky, milani2020intelligent, chen2024integrating, ropelato2018adaptive}. Beyond human learning, some researchers applied concepts akin to ZPD to curriculum design for training autonomous agents ~\cite{seita2019zpd, wang2022zone, tio2023training, tzannetos2024proximal}. However, these modern interpretations of ZPD fail to factor in \emph{assistance} from a more knowledgable teacher, which deviates from the original definition by Vygostky~\cite{vygotsky1978mind}. By contrast, our work explicitly leverages an autonomous agent that shares control with the human learner as the form of assistance for identifying their ZPD.




\paragraph{Skill Modeling}
Skill decomposition has been used for a variety of reasons in robotic planning and reinforcement learning \cite{sutton1999between,andreas2017sketches,shiarlis2018taco,fu2024language}. Likewise, skill decomposition has been show to be an important part of designing  training curriculum for human students in both traditional and, recently, AI-assisted teaching settings \cite{Magill2020-yz, srivastava2022motor}. With \texttt{Z-COACH}, we expand on this line of work to show how to decompose an arbitrary motor control task into a set of human-interpretable skills from noisy auxiliary information (e.g. noisy natural language captions from experts), which we select from when providing personalized instruction.


\section{Formalism}

\begin{figure*} 
\centerline{\includegraphics[width=0.8\textwidth]{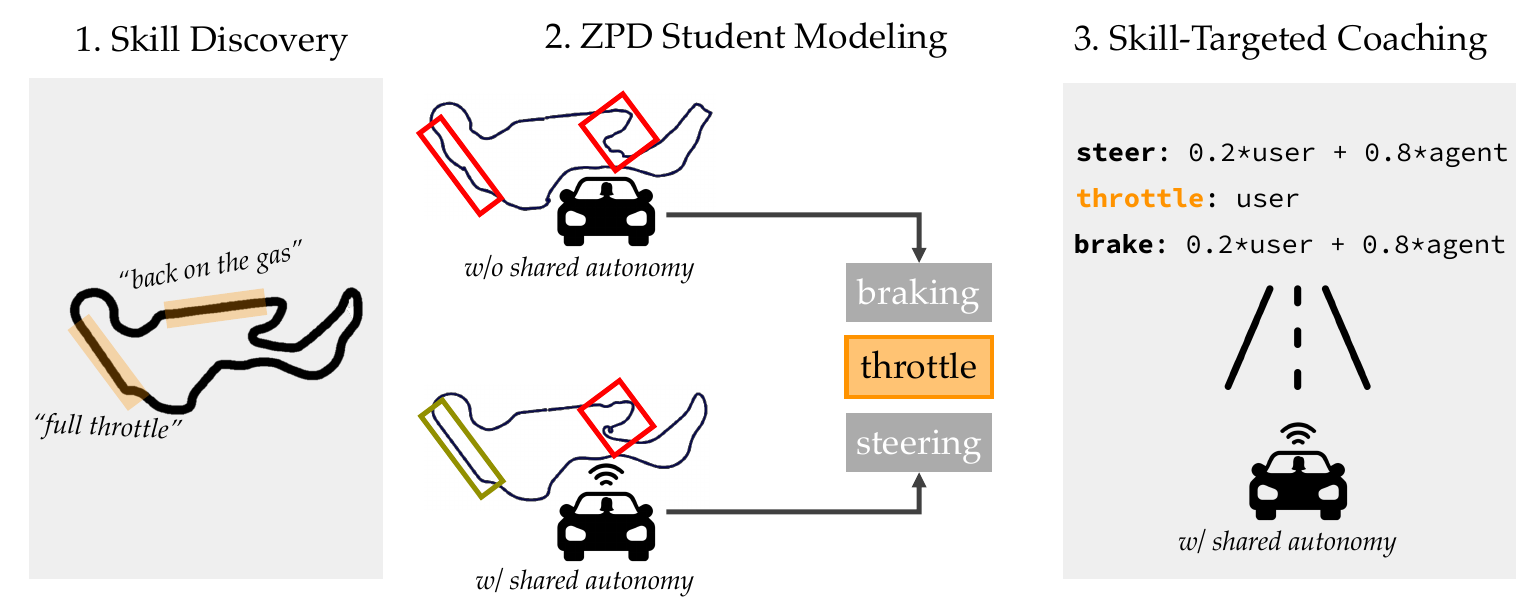}}
\caption{Overview of \texttt{Z-COACH}, which consists of three stages: (1) an interpretable skill discovery stage to identify task-relevant skills to guide coaching, (2) student modeling, which leverages shared autonomy to identify how a driver's behavior changes with assistance in order to choose skills within their ``zone of proximal development'', and (3) skill-targeted coaching, which leverages a different form of shared autonomy that forces drivers to control and practice a specific skill. Note that shared autonomy is leveraged twice by \texttt{Z-COACH}: for student modeling and for coaching.}
\label{fig:zcoach}
\end{figure*}

We now formalize our approach to leveraging shared autonomy for proximal teaching, where our goal is to select optimal coaching interventions that can help guide a student towards optimal behavior for a given task $g$ (e.g. driving). Let us treat $g$ as a standard Markov decision process (MDP) $<\mathcal{S}, \mathcal{A}, f, \mathcal{R}, T>$ with finite horizon $T$, reward function $\mathcal{R}: \mathcal{S} \times \mathcal{A} \rightarrow \mathbb{R}$ 
over state $\mathcal{S}$ and action $\mathcal{A}$ spaces, and a deterministic transition function $f_g: \mathcal{S} \times \mathcal{A} \rightarrow \mathcal{S}$ that maps a state and action pair $s_t, a_t$ at time step $t$ to a new state $s_{t+1}$. We can then define a trajectory $\tau$ as a sequence of state and action pairs $\{s_1,a_1, \dots, s_T,a_T\}$, and can collect trajectories from the student's policy ($\pi_{student}:\mathcal{S} \rightarrow \mathcal{A}$). 

Next, let $\mathcal{Z}_g$ represent the set of task-specific skills (e.g. soft braking), and $\mathcal{C}_g$ be the set of possible coaching interventions. These can include verbal feedback, haptic guidance, or, as in our work, different shared autonomy controls, and one can consider the function $\phi:\mathcal{Z}_g \rightarrow \mathcal{C}_g$ that outputs a specific coaching intervention $c=\phi(z)$ that targets a specific skill $z \in \mathcal{Z}_g$. We therefore seek to learn a useful teaching policy $\pi_{coach}:\mathcal{S},\Pi  \rightarrow \mathcal{C}_g$ that takes a student's current state $s$ and policy $\pi_{student} \sim \Pi$, and selects an intervention $c$ that guides $\pi_{student}$ towards an optimal policy $\pi*$. 

Our primary challenge lies with modeling the transition function $f_{student}:\mathcal{S}, \Pi, \mathcal{C} \rightarrow \Pi$ (i.e., how the coaching  intervention $c$ will modify the student's behavior).  Prior works have tackled this by making assumptions such as modeling student learning over time as a Wiener process, and selecting the coaching intervention $c$ that leads to the biggest increase in proficiency \citep{yu2023coach, ekanadham2017tskirt}. In contrast, we propose leveraging shared autonomy to directly observe how a student's current behavior is affected by assistance. The change in performance can inform which skills $z \in \mathcal{Z}_g$ are most within the student's ``zone of proximal development''  in real-time, allowing us to provide an effective intervention $c=\phi(z)$.

Figure \ref{fig:zcoach} provides an overview of \texttt{Z-COACH}, highlighting how we leverage shared autonomy for both student modeling and coaching. \texttt{Z-COACH} requires three components, which we formalize in more detail below:
\begin{itemize}
    \item Section \ref{sec:sa}: A shared autonomy paradigm that produces a policy $\pi
    _{SA}$ given the student's policy $\pi_{student}$. 
    \item Section \ref{sec:skill}: A set of  interpretable task skills $\mathcal{Z}_g$. 
    \item Section \ref{sec:zpd}: A model \textsf{zpd} which estimates how well each skill $z$ is within the student's ``zone of proximal development'', given $\mathcal{Z}_g$ and $\pi_{SA}$. 
\end{itemize}

\subsection{Shared Autonomy Design} \label{sec:sa}
We consider the simplest form of shared control, where the actions from both the human student ($\pi_{student}$) and an intelligent agent ($\pi_{agent}$) are merged, which in driving has been shown to improve overall performance in safety-critical situations \citep{wang2020review}. Concretely, given an observed state $s$, the output action of the policy $\pi_{SA}$ is:
\begin{equation}\label{eq:sa}
    \pi_{SA}(s) = \alpha * \pi_{agent}(s) + (1 - \alpha) * \pi_{student}(s)
\end{equation}

The term $\alpha$ controls the strength of the autonomy, and is action-specific (e.g. separate $\alpha$ for steering input and brake input in driving) Finally, the autonomous agent $\pi_{agent}$ can come from any source, including planning, reinforcement learning or imitation-learning on expert demonstrations. In our work, we use a PID controller and path planner based on expert trajectories, which we describe in more detail in Section \ref{sec:study}.

\subsection{Interpretable Skill Discovery}
\label{sec:skill}
Recall our goal to model students and design coaching interventions with respect to a set of skills $\mathcal{Z}_g$. Most works in the literature consider hand-crafted skills, and are therefore limited to environments where it is easy to manually evaluate skill performance from a student trajectory $\tau_{student}$, which is tricky for motor control tasks. To address this, \cite{srivastava2022motor} recently proposed using CompILE, an unsupervised skill discovery algorithm, which learns the latent skill set $\mathcal{Z}_g$ from expert trajectories $\tau_{expert} \sim \pi_{expert}$ \cite{kipf2019compile}. Concretely, CompILE jointly trains an encoder $q_\phi$ and decoder $p_\theta$ to segment a trajectory into $M$ segments (marked by boundaries $b$), and represent each segment with latent skill $z \in \mathcal{Z}_g$ that minimizes the following reconstruction loss (black only):

\begin{equation}\label{eq:compile}
    \mathcal{L} = - \mathbb{E}_{q_\theta(b, z|a, s)}\sum_i^M[Pr[\text{seg}_i] * \text{log}p_\theta(a\color{blue}, \psi(s)\color{black}|s, z_i)]
\end{equation}

The output of CompILE can therefore be viewed as a list of $M$ segments, as well as a probability distribution across all skills $z \in \mathcal{Z}_g$ such that $p_{z,i}$ is the likelihood $\text{seg}_i \in M$ expresses skill $z$.
Unfortunately, such a method lacks interpretability, and segments with similar latent skill representations CompILE may not appear similar to humans, making it difficult to map task skills to effective and meaningful coaching interventions ($\phi : \mathcal{Z}_g \rightarrow \mathcal{C}_g)$. We therefore propose using weak supervision by forcing latent skills $z$ to also represent semantically-meaningful auxiliary information (\color{blue}blue \color{black} in Equation \ref{eq:compile}) about the current state. In both Section \ref{results:skills} and Figure \ref{fig:method}, we show how a small amount of language feedback on student driving trajectories helps steer CompILE towards learning more human-interpretable skills $\mathcal{Z}_g$, while retaining the flexibility of not requiring manual evaluation from an expert.
\subsection{Zone of Proximal Development Estimation} \label{sec:zpd}
Finally, we formalize how to use a given shared autonomy policy $\pi_{SA}$ and task skills $\mathcal{Z}_g$ to estimate the student's Zone of Proximal Development.  Recall that our key idea is that  improvement in performance between $\pi_{SA}$ and $\pi_{student}$, with respect to skills $z \in \mathcal{Z}_g$, highlights which skills are most likely to be immediately ``learnable'' for the student.  If we define a function $\mathsf{align}$ that finds the closest (by location) part of a trajectory $\tau$ to a given skill segment $\text{seg}_i \in M$, then we can define:

\begin{multline}\label{eq:zpd}
    \mathsf{zpd}(z) = \sum_{i=0}^{M}p_{z,i}*(\mathsf{score}(\mathsf{align}(\tau_{SA}, \text{seg}_i)) \\ -\mathsf{score}(\mathsf{align}(\tau_{student}, \text{seg}_i)))
\end{multline}

The $\mathsf{score}$ function can be any task-dependent performance metric, such as speed, smoothness, or similiarity to a reference expert trajectory $\tau_{expert}$. Intuitively, $\mathsf{zpd}(z)$ re-weights the difference between a student's performance with and without assistance for a given segment by the relevance of skill $z$. Then, one could define $\phi(z)$ using these weights, such as choosing the intervention $c \in \mathcal{C}_g$ that best targets the skill $\mathsf{argmax}_z\mathsf{zpd}(z)$. Thus, by explicitly incorporating full information about the student's trajectory with assistance $\tau_{SA}$, we can more precisely consider whether a student's difficulty with skill $z$ can be addressed via coaching at the current point of the student's learning.

\section{Overview of Performance Racing Task} \label{sec:study}
\begin{figure*} 
\centerline{\includegraphics[width=0.8\textwidth]{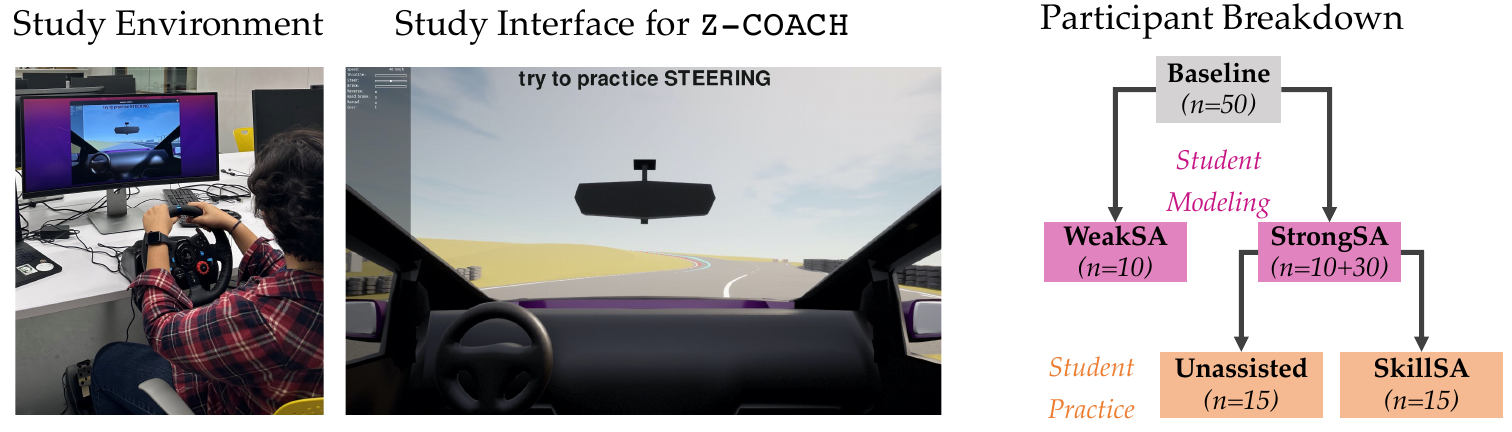}}
\caption{(Left) Image of our study environment for the High Performance Racing task in simulation. We use the CARLA simulator, with an external Logitech G29 steering wheel and pedals. (Middle) Image of the study interface used during coaching. Participants are provided control input information on a sidebar, given verbal guidance on which skill $z$ they should practice, and drive on the track from ego view. (Right) Participant breakdown of our user study. }
\label{fig:method}
\end{figure*}

Although \texttt{Z-COACH} can be applied to learning any motor task where shared autonomy control is feasible, we focus on coaching high-performance racing (HPR), due to the high likelihood of finding novice students and the availability of resources to help evaluate \texttt{Z-COACH}.  Using the open-source CARLA simulator \cite{dosovitskiy2017carla}, specifically designed for autonomous driving research, we simulate the 2-mile Thunderhill West track, part of Thunderhill Raceway Park in California, the venue for the longest automobile race in the United States (See Figure \ref{fig:pull} for example race lines on the race track).

\subsection{Task Description and Environment}
Our goal is to use  \texttt{Z-COACH} to improve novice students' performance when driving one lap around the Thunderhill West track. Strong performance in HPR does not only include a low elapsed time, but also measures such as staying on track and smoothness. Furthermore, the fastest race line around a given track is not necessarily the shortest distance path, particularly in the presence of turns and elevation changes. As such, HPR challenges drivers to perform a diverse set of skills, ranging from properly navigating a hairpin curve to simply maintaining high speed for novices. 

We provide images of our overall task environment and the simulation of the Thunderhill West track in Figure \ref{fig:method}. We modify the default vehicle physics control of the Toyota Prius provided by CARLA, including the torque curve, maximum revolutions per minute, and center of mass. Finally, our environment includes a Logitech G29 Driving Force steering wheel with pedals. 

\subsection{Shared Autonomy Design}
Recall that we use shared autonomy in \texttt{Z-COACH} for both student modeling and coaching. To create the autonomous agent $\pi_{agent}$ used in all shared autonomy modes, we first collected 5 ``expert'' demonstrations from a member of a local HPR club who is familiar with the Thunderhill track, and has extensive driving experience (including over 100 hours in simulation). When a novice student drives, at every time step we first find the nearest point (Euclidean distance over coordinates) in the expert trajectories. We then select a future state at a fixed interval (500 steps) in the expert's trajectory, and use that as the desination for a built-in Planner from CARLA. The planner generates waypoints for use with a PID controller given the current student's state $s$. We use the control outputs as actions for $\pi_{agent}$.

For student modeling, we consider two shared autonomy modes:
\begin{itemize}
    \item \textsc{StrongSA}: Following Equation \ref{eq:sa}, where $\alpha=0.8$.
    \item \textsc{WeakSA}: Following Equation \ref{eq:sa}, where $\alpha=0.05$.
\end{itemize}

We broadly refer to the skill-focused shared autonomy used in coaching as \textsc{SkillSA}, though in practice \textsc{SkillSA} follows Equation \ref{eq:sa} and sets $\alpha=0.8$ for all controls \textit{except} one of Throttle, Brake, or Steering, for which $\alpha=0$, depending on the skill identified to practice.

\begin{figure} 
\centerline{\includegraphics[width=0.3\textwidth]{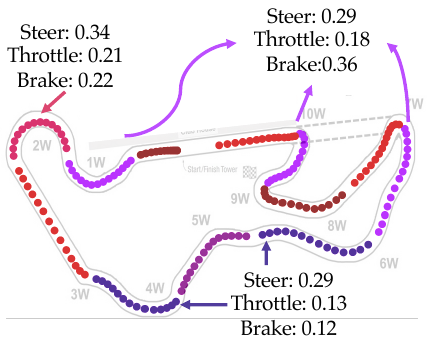}}
\caption{Output segmentations produced by CompILE over an expert driver's trajectory around the Thunderhill West race track. By modifying CompILE to take as input noisy language annotations, the resulting segmentation is more interpretable and aligned with human notions of skills. }
\label{fig:skill}
\end{figure}

For all shared autonomy modes, if the driver goes off track above a threshold (e.g. into the grass hills in the simulation environment), we revert to full student control. Nevertheless, there is an important distinction between the shared autonomy used for student modeling (\textsc{StrongSA}, \textsc{WeakSA}) versus coaching (\textsc{SkillSA}), as the latter forces to the driver to focus on providing full control in one dimension (the skill recommended for practice by \texttt{Z-COACH})  while simultaneously adapting to the assistance along other skill dimensions. 

\subsection{Human Subject Study} We recruited 50 participants for a human subject study to evaluate \texttt{Z-COACH} (see Figure \ref{fig:method} for breakdown). The study aimed to observe improvements in driving performance over time, with a particular focus on lap completion, time, and avoiding crashes. Participants were majority university students (92\%) with no driving experience on a race track  (100\%), but had a wide range of experience with driving (0-30 hrs/week) and playing video games (0-10 hrs/week). We did not collect any further demographic information, and the study was approved by our Institutional Review Board.

The entire study consisted of the following stages:
\begin{enumerate}
    \item \textbf{Baseline:} 2 x Unassisted Trials  
    \item 2 x Assisted Trials ( \textsc{StrongSA} or \textsc{WeakSA}).
    \item 1 x Unassisted Trial  
    \item 5 Minutes of Practice (Unassisted or \textsc{SkillSA})
    \item \textbf{Evaluation:}  3 x Unassisted Trial
\end{enumerate}

Apart from the 5-minute practice stage, the  subject's goal in each trial was to complete 1 lap of the race track before 3 minutes elapsed. During the practice session, participants had full control over how they approached the task and could reset to the starting point as needed. Throughout the study, subjects were presented with a visual guide showing the optimal path of an expert driver. Finally, for all participants, we recorded their driving trajectory starting from their first control input, and user interpolation was applied to normalize each trajectory to sample data every second.

We randomly assigned the first 20 participants to \textsc{StrongSA} or \textsc{WeakSA} for Stage 2, which, along with the trials in Stage 1 and 3, was used for student modeling following Equation 
\ref{eq:zpd}. As we will describe in Section \ref{sec:empzpd}, we empirically found that \textsc{StrongSA} led to stronger student modeling more aligned with an expert coach. The remaining 30 participants were therefore assigned to \textsc{StrongSA} for Stage 2, and randomly assigned to receive no assistance or \textsc{SkillSA} in Stage 4. The particular coaching intervention that \textsc{SkillSA} uses for each participant is based on $\mathsf{argmax}_{z \in (steer, brake, throttle)}\textsf{zpd}(z)$, or the skill that received the highest score in our student modeling based on that participant's trajectories in Stages 1-3. Note that while we set the size of set $\mathcal{Z}_g=8$, we restrict coach actions to only consider steer, brake, and throttle. 

Upon completing the study, all participants were asked to fill out a feedback form, where they reflected on the effectiveness of the five-minute practice session, and provided additional feedback on their experience with the simulator and assistance. We provide more details, including the full set of instructions participants received, in the Appendix. 

Overall, the structure of the study allowed us to assess the influence of shared autonomy on learning  by comparing each participant's \textbf{Baseline} and \textbf{Evaluation} rounds, as well as systematically evaluate each component of \texttt{Z-COACH}.


\section{Experimental Results}




\begin{figure*}
\centerline{\includegraphics[width=0.95\textwidth]{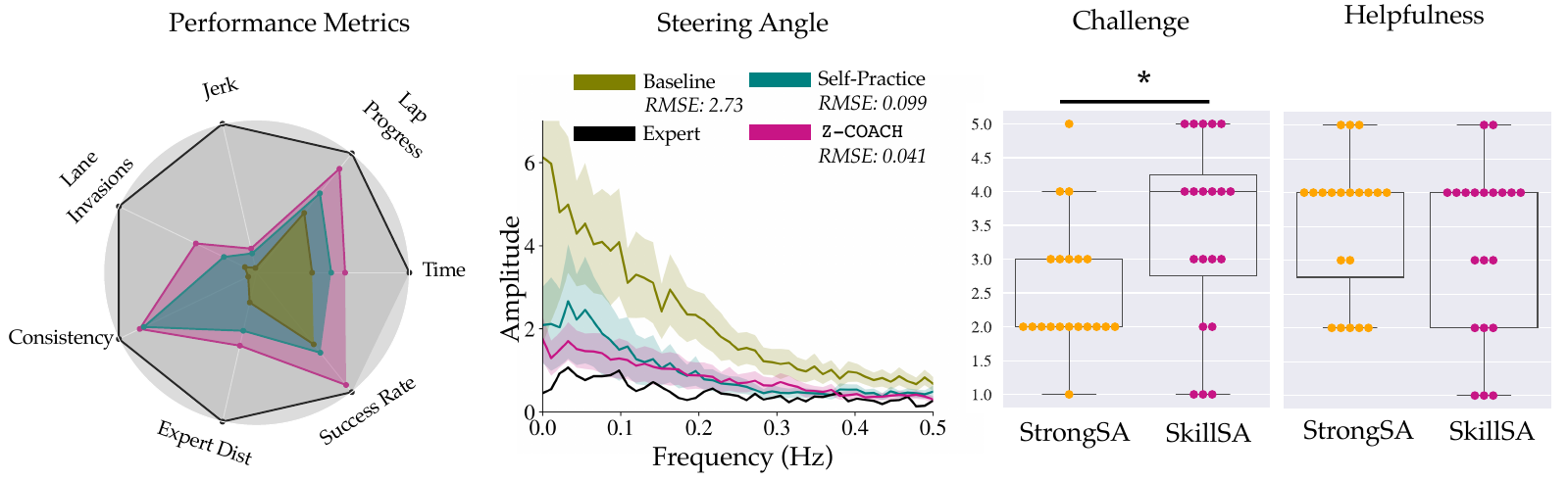}}
\caption{ (Left) Comparison of overall performance between Baseline trials, Expert trials, and Evaluation trials from participants either assigned Self-Practice or receiving assistance via \texttt{Z-COACH}, across a variety of quantitative metrics highlights improvement from \texttt{Z-COACH}. (Middle) Fourier transform signal analysis plot of the steering wheel input shows that \texttt{Z-COACH} guides students towards similar steering behavior as an expert HPR driver, including decreasing the amount of large, low frequency turns. (Right) Participant feedback shows that \textsc{SkillSA}, used for coaching, is found challenge participants  significantly more  than \textsc{StrongSA}, yet comparable in helpfulness. * marks statistical significance ($p < 0.05$ with a paired t-test.) }
\label{fig:results}
\end{figure*}

Recall that \texttt{Z-COACH} consists of three steps: (i) task skill discovery, (ii) student modeling with shared autonomy to estimate how much a skill is within a student's ``zone of proximal development'', and (iii)  using skill-focused shared autonomy to help the student improve. We now evaluate each step. 
\subsection{Interpretable Skill Discovery}\label{results:skills} As described in Section \ref{sec:skill}, we wish to identify the different skills required in our HPR task in order to provide effective coaching interventions. Unfortunately, simply running an unsupervised skill discovery algorithm may return segments corresponding to latent skills $z \in \mathcal{Z}_g$, but we have no way of interpreting what those skills represent, making it tricky to provide focused skill-coaching via shared autonomy. 

We therefore augment our dataset of expert trajectories with noisy language annotations, a form of auxiliary information as in Equation \ref{eq:compile}. Weak supervision in the form of language describing actions appears in a variety of places, including coaching videos on YouTube, or user-generated descriptions on traffic applications. For our task, we use a small subset of data collected and shared by the Toyota Research Institute. Their dataset consists of survey and driving data from 15 participants having one-on-one simulator performance driving training from 1 racing coach on the Thunderhill race track, and includes language feedback such as \textit{``straighten the wheel''} or \textit{``off the gas here''} that corresponds to a particular $(s,a)$ in a student's trajectory. We use data from only \textit{one} participant, and provide further information about this dataset in the Appendix.

Our first step is to cluster the noisy language annotations into semantically meaningful groups, so that we can use the cluster index as the auxiliary information in Equation \ref{eq:compile}. We use the llama-3 language model to generate cluster mappings and descriptions by providing the following input string: 
\begin{quote}
\small
\texttt{You will be given a list of feedback to a driving student. Please cluster them into N skills. The driver can control a throttle, steering wheel, and brake. Return a dictionary that maps each string to a cluster ID, and another dictionary that maps  the cluster ID to its description.}  
\end{quote}

We set $N=8$, and show in Table \ref{tab:clusters} the output descriptions and example feedback for each cluster. Several of the clusters (e.g. Braking, Throttle, Steering) are well-aligned with the task action space $\mathcal{A}$, making it feasible to consider shared autonomy for skill-focused coaching. 

However, in order to identify \textit{which} skill is appropriate for a student to learn, we need to map these clusters to trajectory segments. Following the approach described in Section \ref{sec:skill}, we train CompILE on expert trajectories with the modified loss function in \ref{eq:compile}, again using $N=8$ as a hyperparameter, and let the auxiliary information $\psi(s)$ equal the cluster ID (e.g. 6) of the language feedback provided at the state closest to $s$ in position. The resulting skill segmentation, shown in Figure \ref{fig:skill}, successfully groups together parts of the track that requires high throttle without any steering, as well as the entry into sharp turns. Furthermore, when compared with training CompILE without weak supervision, our approach significantly improves the compression ratio ($\textbf{1.83}$ vs. $\textbf{1.22}$) of the resulting segmentation, with little impact on Mean Square Error ($\textbf{0.037}$ vs. $\textbf{0.037}$) of the reconstruction. Finally, we can identify which Cluster ID, and therefore language feedback, is most associated with a given latent $z$ by examining the output logits of the decoder $p_\theta$  in CompILE (see Section \ref{sec:skill}). 

Overall, our results show success in identifying semantically meaningful skills in task trajectories, which we will use for student modeling.

\begin{table}[htbp]\label{tab:clusters}
\caption{Unsupervised Clustering of Verbal Feedback}
\begin{center}
\begin{tabular}{cc}
\textbf{Cluster Description}&  \textbf{Example Feedback}  \\
 \hline
Braking & \textit{hold the brake hard} \\
Lane positioning & \textit{stay to the right}\\
Steering control & \textit{little bit of steering} \\
Car handling  & \textit{let the car push out} \\
Throttle control & \textit{now squeeze the gas}\\
Navigation &  \textit{up the hill}\\
Target aiming & \textit{aim for the right} \\
Encouragement & \textit{there you go} \\

\end{tabular}
\label{tab1}
\end{center}
\end{table}

\begin{figure}
\centerline{\includegraphics[width=0.35\textwidth]{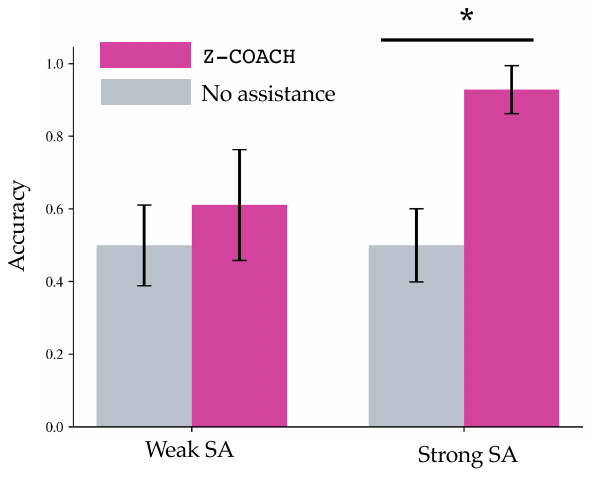}}
\caption{For participants assigned either Weak and Strong Shared Autonomy during the Student Modeling stage, incorporating their performance and behavior when driving with assistance led to higher accuracy in predicting which skill a certain driver should be focused on learning next. For ground truth labels, we showed videos of the student driver trajectories to a professional HPR coach, who provided ranked ordering of skills per participant.  * denotes statistical significance ($p < 0.05$) using Welch's t-test.}
\label{fig:zpd}
\end{figure}

\subsection{Estimating Students' Zone of Proximal Development} \label{sec:empzpd}

We next evaluate the benefit of using shared autonomy for student modeling. We screen captured videos of 20 participants, split evenly between receiving \textsc{StrongSA} or \textsc{WeakSA} in Stage 2 of our study, and hired a professional high performance racing coach (compensated 75\$ per hour) to watch each video and provide ground truth rankings for the order in which he would coach the student to focus on the following three skills: Steering, Braking, Throttle. While there was variation between participants, the coach shared that most participants needed to improve their Throttle control before even trying to practice Braking, indicating that the Braking skill is less likely than Throttle to be within a student's ``Zone of Proximal Development''. Indeed, only two participants were labeled as needing to first focus on practicing braking.

We next analyze rankings produced by $\mathsf{zpd}(z)$ for each of the 8 possible $z \in \mathcal{Z}_g$. We average all assisted trajectories $\tau_{SA}$ from Stage 2 and baseline trajectories $\tau_{student}$ from Stage 1 before calculating  $\mathsf{zpd}(z)$, using the Dynamic Time Warp distance between trajectories from the student and our reference expert as $\textsf{score}$. We find that participants assigned \textsc{StrongSA} in Stage 2 had more than 90\% accuracy when comparing with the ground truth labels from the coach (See Figure \ref{fig:zpd}). Meanwhile, when $\mathsf{zpd}(z)$ is calculated without a term capturing the student's performance with assistance (i.e. replacing $\mathsf{score}(\mathsf{align}(\tau_{SA}, \text{seg}_i))$ with a constant in Equation \ref{eq:zpd}), the accuracy significantly degrades. Surprisingly, the benefit of measuring $\mathsf{zpd}(z)$ with  \textsc{WeakSA} assistance is not as strong, despite using more student inputs. This might be due to user frustration with less helpful assistance affecting their performance. Nevertheless, incorporating student performance with \textsc{WeakSA} assistance when calculating $\mathsf{zpd}(z)$ still improves accuracy over the constant case, and our overall results confirm that shared autonomy does indeed lead to a stronger model of what skills a student should next focus on learning.

\begin{table}[htbp]\label{tab:metrics}
\caption{Mean Change Between Baseline and Evaluation Trials}
\begin{center}
\begin{tabular}{ccc}
\textbf{Metric}&  \textbf{Self-Practice}  &  \texttt{\textbf{Z-COACH}} \\
\hline
$\Delta$ Success Rate $\uparrow$  & 0.07 [-0.20, 0.33] & 0.28 [0.0632, 0.4924] \\
$\Delta$ Lap Progress $\uparrow$  & 0.22 [-0.35, 0.80] & 0.69 [0.08, 1.29] \\
$\Delta$ Lap Time (s) $\downarrow$  & -11 [-23.8, 2.6]*
  & \textbf{-29 [-38.7, -19.5] *} \\
$\Delta$ Consistency $\uparrow$ & 2.6 [-2.18, 7.38] & 2.4 [-0.19, 5.02] \\ 
$\Delta$ Expert Distance $\downarrow$&  -1.11 [-2.47, 0.33] & -5.39 [-13.48, 2.70] \\
$\Delta$ Jerk $\downarrow$ & 2.1 [-3.51, 7.85] * & \textbf{-6.6  [-12.16, -0.99] *} \\ 
$\Delta$ $\#$ Lane Invasions $\downarrow$ & -3.4 [-21.2, 14.4] * & \textbf{-33.5 [-49.1, -17.9] *} \\ 
\end{tabular}
\footnotesize{* denotes a statistical significant difference between mean change of Self-Practice and \texttt{Z-COACH} conditions ($p < 0.05$, using Welch's t-test)}
\label{tab1}
\end{center}
\end{table} 

\subsection{Improving Student Learning}
Finally, we seek to evaluate whether the skill-focused coaching aspect of \texttt{Z-COACH} does indeed help improve student performance. Recall that participants were randomly selected to either practice independently for 5-minutes, or provided \textsc{SkillSA} assistance which forces drivers to take full control on either Steering, Throttle, or Brake, depending on their skill level. Figure \ref{fig:method} shows an example coaching interface for a participant.

Between both groups (self-practice or \textsc{SkillSA}), we compare each student's performance change between the Stage 5 evaluation trials and the Stage 1 baseline trials, with respect to the following evaluation metrics:
\begin{itemize}
    \item \textbf{Success Rate}: Average \% of times the driver completed a full lap under the 3 minutes time-limit
    \item \textbf{Lap Progress}: Proportion of the lap covered
    \item \textbf{Lap Time}: Time taken if the driver completed the full track
    \item \textbf{Expert Distance}: Distance between the student and expert trajectories, measured with Dynamic Time Warp \citep{giorgino2009dtw}
    \item \textbf{Consistency}: The mean distance between trajectories for each pair of trials within the same stage, measured with Dynamic Time Warp \citep{giorgino2009dtw} 
    \item \textbf{Jerk}: Change in magnitude of jerk, or the rate of change in acceleration over time, a measure of trajectory smoothness
    \item \textbf{\# Lane Invasions}: Number of times the vehicle crosses lane lines and goes off track
\end{itemize}

Table \ref{tab:metrics} shows that across all metrics, students provided skill-focused coaching with \texttt{Z-COACH} not only improved over time, but also by  a stronger amount than students who had to self-practice (except for $\Delta$ Consistency). Using a Welch's t-test to test for statistical significance at $p < 0.05$ with Bonferroni correction, we find significant effects with \texttt{Z-COACH} for improvement in lap time, jerk, and average number of lane invasions, suggest that \texttt{Z-COACH} helps students drive faster, more smoothly, and with more control over staying on the race track. 
Beyond improving student performance, Figure \ref{fig:results} shows the degree to which \texttt{Z-COACH} reaches expert performance across different metrics, highlighting that there is still room for further improvement from coaching for smoothness (jerk) and reducing the amount of lane invasions. 

We further analyze how \texttt{Z-COACH} affects student steering behavior by providing a Fourier analysis of steering angle in Figure \ref{fig:results}. We observed during the user study that novices often took large, ``swinging'' turns, which corresponds to the high amplitude and low frequency values for the Baseline trials. While self-practice does reduce such behavior, the curve for \texttt{Z-COACH} has the lowest RMSE when compared with the expert, suggesting that \texttt{Z-COACH} best leads students towards optimal steering behavior. 

\subsection{Student Feedback}
While \textsc{SkillSA} leads to stronger learning outcomes compared to self-practice, we were interested in how participants viewed skill-focused shared autonomy with the \textsc{StrongSA} assistance they received during the student modeling stage of their study. We asked all participants to rate (range 1 to 5, with 5 indicating agreement) how \textit{helpful} they found the assistance, and to what degree the assistance \textit{challenged} them. As Figure \ref{fig:results} shows, while there was no significant difference in how helpful participants perceived both assistant types, they found that \textsc{SkillSA} significantly challenged them more (significance at  $p<0.05$ determined with a paired t-test). This observation is further supported by their free-text responses; for example, one participant described the \textsc{StrongSA} assistance received during  student modeling as \textit{``it was hard for me to know how to improve my driving skills"}, but then described the \textsc{SkillSA} assistance it received during coaching as \textit{``This more interactive mode helps me more to get a sense when I should adjust my previous strategy on throttle''}. 

Meanwhile, another participant who received \textsc{SkillSA} assistance targeting steering shared that \textit{``I had a lot of difficulty, and  didn't feel like I was better when doing it. But afterwards in the evaluation trials I swear I had improved my steering''}. These results raise the interesting question of whether coaching interventions that help a student learn are actually perceived as helpful at the time, and what properties to consider when developing shared autonomy assistance intended to help students learn.

\section{Limitations \& Conclusion}
We have shown that \texttt{Z-COACH} effectively applies shared autonomy to both improve student modeling \textit{and} provide skill-targeted coaching, leading to higher overall improvement in performance compared to self-practice. While we focused on high performance racing due to the availability of coaching data and simulation environment, as well as its novelty for participants, an important next step is to evaluate what properties of different motor control tasks may affect the ability of  \texttt{Z-COACH} to improve student learning. Furthermore, there exist many confounders (e.g. frame rate of simulator, lack of haptic feedback) that could affect results. Finally, while we intentionally chose simpler forms of shared autonomy to increase focus on our core task of motor learning, future work could explore more complex forms of shared autonomy, such as incorporating agent confidence. Our work can easily be integrated with other forms of teaching (e.g. with haptics, verbal feedback), where \texttt{Z-COACH} would be used to identify which skills the instruction should target. Overall, \texttt{Z-COACH} is a first step towards re-imagining the role of shared autonomy to not only assist humans in performing tasks, but to enable effective and specialized skill learning. 

\section{Acknowledgements}
We thank Jack Bernardo for help with collecting expert demonstrations, Ngorli Paintsil for support with the early stages of this project, and all user study participants. This work was supported by the Toyota Research Institute (TRI).


\footnotesize
\bibliographystyle{abbrvnat}
\balance
\bibliography{main}

\begin{thebibliography}{57}
\providecommand{\natexlab}[1]{#1}
\providecommand{\url}[1]{\texttt{#1}}
\expandafter\ifx\csname urlstyle\endcsname\relax
  \providecommand{\doi}[1]{doi: #1}\else
  \providecommand{\doi}{doi: \begingroup \urlstyle{rm}\Url}\fi

\bibitem[Abbink et~al.(2018)Abbink, Carlson, Mulder, De~Winter, Aminravan, Gibo, and Boer]{abbink2018topology}
D.~A. Abbink, T.~Carlson, M.~Mulder, J.~C. De~Winter, F.~Aminravan, T.~L. Gibo, and E.~R. Boer.
\newblock A topology of shared control systems—finding common ground in diversity.
\newblock \emph{IEEE Transactions on Human-Machine Systems}, 48\penalty0 (5):\penalty0 509--525, 2018.

\bibitem[Aigner and McCarragher(1997)]{aigner1997shared}
P.~Aigner and B.~McCarragher.
\newblock Human integration into robot control utilising potential fields.
\newblock In \emph{Proceedings of International Conference on Robotics and Automation}, volume~1, pages 291--296 vol.1, 1997.
\newblock \doi{10.1109/ROBOT.1997.620053}.

\bibitem[Andreas et~al.(2017)Andreas, Klein, and Levine]{andreas2017sketches}
J.~Andreas, D.~Klein, and S.~Levine.
\newblock Modular multitask reinforcement learning with policy sketches.
\newblock In \emph{International Conference on Machine Learning (ICML)}, 2017.

\bibitem[Argall(2018)]{argall2018autonomy}
B.~D. Argall.
\newblock Autonomy in rehabilitation robotics: An intersection.
\newblock \emph{Annual Review of Control, Robotics, and Autonomous Systems}, 1\penalty0 (1):\penalty0 441--463, 2018.

\bibitem[Bastani et~al.(2024)Bastani, Bastani, Sungu, Ge, Kabakcı, and Mariman]{bastani2024generative}
H.~Bastani, O.~Bastani, A.~Sungu, H.~Ge, O.~Kabakcı, and R.~Mariman.
\newblock Generative ai can harm learning.
\newblock \emph{The Wharton School Research Paper}, 2024.
\newblock URL \url{https://ssrn.com/abstract=4895486}.
\newblock Accessed: 2025-01-04.

\bibitem[Bobu et~al.(2020)Bobu, Scobee, Fisac, Sastry, and Dragan]{bobu2020less}
A.~Bobu, D.~R.~R. Scobee, J.~F. Fisac, S.~S. Sastry, and A.~D. Dragan.
\newblock Less is more: Rethinking probabilistic models of human behavior.
\newblock In \emph{Proceedings of the 2020 ACM/IEEE International Conference on Human-Robot Interaction}, HRI '20, page 429–437, New York, NY, USA, 2020. Association for Computing Machinery.
\newblock ISBN 9781450367462.
\newblock \doi{10.1145/3319502.3374811}.
\newblock URL \url{https://doi.org/10.1145/3319502.3374811}.

\bibitem[Bragg and Brunskill(2020)]{bragg2020fake}
J.~Bragg and E.~Brunskill.
\newblock Fake it till you make it: Learning-compatible performance support.
\newblock In R.~P. Adams and V.~Gogate, editors, \emph{Proceedings of The 35th Uncertainty in Artificial Intelligence Conference}, volume 115 of \emph{Proceedings of Machine Learning Research}, pages 915--924. PMLR, 22--25 Jul 2020.
\newblock URL \url{https://proceedings.mlr.press/v115/bragg20a.html}.

\bibitem[Byeon et~al.(2024)Byeon, Choi, Zhang, and Hwang]{byeon2024flight}
S.~Byeon, J.~Choi, Y.~Zhang, and I.~Hwang.
\newblock Stochastic-skill-level-based shared control for human training in urban air mobility scenario.
\newblock \emph{J. Hum.-Robot Interact.}, 13\penalty0 (3), Aug. 2024.
\newblock \doi{10.1145/3603194}.
\newblock URL \url{https://doi.org/10.1145/3603194}.

\bibitem[Chen et~al.(2024)Chen, Alghowinem, Breazeal, and Park]{chen2024integrating}
H.~Chen, S.~Alghowinem, C.~Breazeal, and H.~W. Park.
\newblock Integrating flow theory and adaptive robot roles: A conceptual model of dynamic robot role adaptation for the enhanced flow experience in long-term multi-person human-robot interactions.
\newblock In \emph{Proceedings of the 2024 ACM/IEEE International Conference on Human-Robot Interaction}, pages 116--126, 2024.

\bibitem[Chen et~al.(2022)Chen, Fong, and Soh]{chen2022mirror}
K.~Chen, J.~Fong, and H.~Soh.
\newblock Mirror: Differentiable deep social projection for assistive human-robot communication.
\newblock 2022.

\bibitem[Clement et~al.(2015)Clement, Roy, Oudeyer, and Lopes]{clement2015multi}
B.~Clement, D.~Roy, P.-Y. Oudeyer, and M.~Lopes.
\newblock Multi-armed bandits for intelligent tutoring systems.
\newblock \emph{Journal of Educational Data Mining}, 7\penalty0 (2), 2015.

\bibitem[Corbett and Anderson(2005)]{corbett2005kt}
A.~T. Corbett and J.~R. Anderson.
\newblock Knowledge tracing: Modeling the acquisition of procedural knowledge.
\newblock \emph{User Modeling and User-Adapted Interaction}, 4:\penalty0 253--278, 2005.
\newblock URL \url{https://api.semanticscholar.org/CorpusID:19228797}.

\bibitem[Csibra and Gergely(2009)]{csibra2009pedagogy}
G.~Csibra and G.~Gergely.
\newblock Natural pedagogy.
\newblock \emph{Trends in Cognitive Sciences}, 13\penalty0 (4):\penalty0 148--153, 2009.
\newblock ISSN 1364-6613.
\newblock \doi{https://doi.org/10.1016/j.tics.2009.01.005}.
\newblock URL \url{https://www.sciencedirect.com/science/article/pii/S1364661309000473}.

\bibitem[Cui et~al.(2023)Cui, Karamcheti, Palleti, Shivakumar, Liang, and Sadigh]{cui2023no}
Y.~Cui, S.~Karamcheti, R.~Palleti, N.~Shivakumar, P.~Liang, and D.~Sadigh.
\newblock No, to the right: Online language corrections for robotic manipulation via shared autonomy.
\newblock In \emph{Proceedings of the 2023 ACM/IEEE International Conference on Human-Robot Interaction}, pages 93--101, 2023.

\bibitem[Da~Lio et~al.(2021)Da~Lio, Don{\`a}, Papini, and Plebe]{da2021biasing}
M.~Da~Lio, R.~Don{\`a}, G.~P.~R. Papini, and A.~Plebe.
\newblock The biasing of action selection produces emergent human-robot interactions in autonomous driving.
\newblock \emph{IEEE Robotics and Automation Letters}, 7\penalty0 (2):\penalty0 1254--1261, 2021.

\bibitem[David et~al.(2016)David, Segal, and Gal]{david2016sequencing}
Y.~B. David, A.~Segal, and Y.~Gal.
\newblock Sequencing educational content in classrooms using bayesian knowledge tracing.
\newblock In \emph{Proceedings of the sixth international conference on Learning Analytics \& Knowledge}, pages 354--363, 2016.

\bibitem[de~Winter et~al.(2023)de~Winter, Petermeijer, and Abbink]{de-Winter2023-cp}
J.~C.~F. de~Winter, S.~M. Petermeijer, and D.~A. Abbink.
\newblock Shared control versus traded control in driving: a debate around automation pitfalls.
\newblock \emph{Ergonomics}, 66\penalty0 (10):\penalty0 1494--1520, Oct. 2023.

\bibitem[Dosovitskiy et~al.(2017)Dosovitskiy, Ros, Codevilla, Lopez, and Koltun]{dosovitskiy2017carla}
A.~Dosovitskiy, G.~Ros, F.~Codevilla, A.~Lopez, and V.~Koltun.
\newblock Carla: An open urban driving simulator.
\newblock \emph{arXiv preprint arXiv:1711.03938}, 2017.

\bibitem[Ekanadham and Karklin(2017)]{ekanadham2017tskirt}
C.~Ekanadham and Y.~Karklin.
\newblock T-skirt: Online estimation of student proficiency in an adaptive learning system, 2017.
\newblock URL \url{https://arxiv.org/abs/1702.04282}.

\bibitem[Fu et~al.(2024)Fu, Sharma, Stengel-Eskin, Konidaris, Roux, C{\^o}t{\'e}, and Yuan]{fu2024language}
H.~Fu, P.~Sharma, E.~Stengel-Eskin, G.~Konidaris, N.~L. Roux, M.-A. C{\^o}t{\'e}, and X.~Yuan.
\newblock Language-guided skill learning with temporal variational inference.
\newblock \emph{arXiv preprint arXiv:2402.16354}, 2024.

\bibitem[Giorgino(2009)]{giorgino2009dtw}
T.~Giorgino.
\newblock omputing and visualizing dynamic time warping alignments in r: The dtw package.
\newblock \emph{Journal of Statistical Software}, 2009.

\bibitem[Gopinath and Argall(2020)]{gopinath2020active}
D.~E. Gopinath and B.~D. Argall.
\newblock Active intent disambiguation for shared control robots.
\newblock \emph{IEEE Transactions on Neural Systems and Rehabilitation Engineering}, 28\penalty0 (6):\penalty0 1497--1506, 2020.

\bibitem[Ho et~al.(2016)Ho, Littman, MacGlashan, Cushman, and Austerweil]{ho2016showing}
M.~K. Ho, M.~Littman, J.~MacGlashan, F.~Cushman, and J.~L. Austerweil.
\newblock Showing versus doing: Teaching by demonstration.
\newblock In D.~Lee, M.~Sugiyama, U.~Luxburg, I.~Guyon, and R.~Garnett, editors, \emph{Advances in Neural Information Processing Systems}, volume~29. Curran Associates, Inc., 2016.
\newblock URL \url{https://proceedings.neurips.cc/paper_files/paper/2016/file/b5488aeff42889188d03c9895255cecc-Paper.pdf}.

\bibitem[Ho et~al.(2018)Ho, Littman, Cushman, and Austerweil]{ho2018effectively}
M.~K. Ho, M.~L. Littman, F.~Cushman, and J.~L. Austerweil.
\newblock Effectively learning from pedagogical demonstrations.
\newblock In \emph{Proceedings of the Annual Meeting of the Cognitive Science Society}, volume~40, 2018.
\newblock URL \url{https://escholarship.org/uc/item/16v54626}.
\newblock Accessed: 2025-01-04.

\bibitem[Hou et~al.(2023)Hou, Yu, Hsu, and Yu]{hou2023teachingbot}
Z.~Hou, C.~Yu, D.~Hsu, and H.~Yu.
\newblock Teachingbot: Robot teacher for human handwriting.
\newblock \emph{arXiv preprint arXiv:2309.11848}, 2023.

\bibitem[Karamcheti et~al.(2022)Karamcheti, Srivastava, Liang, and Sadigh]{karamcheti2022lila}
S.~Karamcheti, M.~Srivastava, P.~Liang, and D.~Sadigh.
\newblock Lila: Language-informed latent actions.
\newblock In \emph{Conference on Robot Learning}, pages 1379--1390. PMLR, 2022.

\bibitem[Kipf et~al.(2019)Kipf, Li, Dai, Zambaldi, Sanchez-Gonzalez, Grefenstette, Kohli, and Battaglia]{kipf2019compile}
T.~Kipf, Y.~Li, H.~Dai, V.~Zambaldi, A.~Sanchez-Gonzalez, E.~Grefenstette, P.~Kohli, and P.~Battaglia.
\newblock Compile: Compositional imitation learning and execution, 2019.
\newblock URL \url{https://arxiv.org/abs/1812.01483}.

\bibitem[Li et~al.(2019)Li, Song, Cao, Wang, Huang, Hu, and Wang]{li2019shared}
M.~Li, X.~Song, H.~Cao, J.~Wang, Y.~Huang, C.~Hu, and H.~Wang.
\newblock Shared control with a novel dynamic authority allocation strategy based on game theory and driving safety field.
\newblock \emph{Mechanical Systems and Signal Processing}, 124:\penalty0 199--216, 2019.

\bibitem[Losey et~al.(2020)Losey, Srinivasan, Mandlekar, Garg, and Sadigh]{losey2020latent}
D.~P. Losey, K.~Srinivasan, A.~Mandlekar, A.~Garg, and D.~Sadigh.
\newblock Controlling assistive robots with learned latent actions.
\newblock In \emph{International Conference on Robotics and Automation (ICRA)}, pages 378--384, 2020.

\bibitem[Luckin and Du~Boulay(1999)]{luckin1999ecolab}
R.~Luckin and B.~Du~Boulay.
\newblock Ecolab: The development and evaluation of a vygotskian design framework.
\newblock \emph{International Journal of Artificial Intelligence in Education}, 10:\penalty0 198--220, 01 1999.

\bibitem[Magill and Anderson(2020)]{Magill2020-yz}
R.~A. Magill and D.~Anderson.
\newblock \emph{Motor learning and control: Concepts and applications}.
\newblock Feb. 2020.

\bibitem[Marcano et~al.(2020)Marcano, D{\'\i}az, P{\'e}rez, and Irigoyen]{marcano2020review}
M.~Marcano, S.~D{\'\i}az, J.~P{\'e}rez, and E.~Irigoyen.
\newblock A review of shared control for automated vehicles: Theory and applications.
\newblock \emph{IEEE Transactions on Human-Machine Systems}, 50\penalty0 (6):\penalty0 475--491, 2020.

\bibitem[Milani et~al.(2020)Milani, Fan, Gulati, Nguyen, Fang, and Yadav]{milani2020intelligent}
S.~Milani, Z.~Fan, S.~Gulati, T.~Nguyen, F.~Fang, and A.~Yadav.
\newblock Intelligent tutoring strategies for students with autism spectrum disorder: A reinforcement learning approach.
\newblock In \emph{The 2020 CMU Symposium on Artificial Intelligence and Social Good}, 2020.

\bibitem[Mower et~al.(2021)Mower, Moura, and Vijayakumar]{mower2021sharedcontrol}
C.~Mower, J.~Moura, and S.~Vijayakumar.
\newblock Skill-based shared control.
\newblock In \emph{Robotics: Science and Systems XVII}. The Robotics: Science and Systems Foundation, July 2021.
\newblock \doi{10.15607/RSS.2021.XVII.028}.
\newblock URL \url{https://roboticsconference.org/}.
\newblock Robotics: Science and Systems 2021, R:SS 2021 ; Conference date: 12-07-2021 Through 16-07-2021.

\bibitem[Nemlekar et~al.(2023)Nemlekar, Dhanaraj, Guan, Gupta, and Nikolaidis]{nemlekar2023robot}
H.~Nemlekar, N.~Dhanaraj, A.~Guan, S.~K. Gupta, and S.~Nikolaidis.
\newblock Transfer learning of human preferences for proactive robot assistance in assembly tasks.
\newblock In \emph{Proceedings of the 2023 ACM/IEEE International Conference on Human-Robot Interaction}, HRI '23, page 575–583, New York, NY, USA, 2023. Association for Computing Machinery.
\newblock ISBN 9781450399647.
\newblock \doi{10.1145/3568162.3576965}.
\newblock URL \url{https://doi.org/10.1145/3568162.3576965}.

\bibitem[Okamura et~al.(2010)Okamura, Matari{\'c}, and Christensen]{okamura2010medical}
A.~M. Okamura, M.~J. Matari{\'c}, and H.~I. Christensen.
\newblock Medical and health-care robotics.
\newblock \emph{IEEE Robotics \& Automation Magazine}, 17\penalty0 (3):\penalty0 26--37, 2010.

\bibitem[Reddy et~al.(2018)Reddy, Dragan, and Levine]{reddy2018shared}
S.~Reddy, A.~D. Dragan, and S.~Levine.
\newblock Shared autonomy via deep reinforcement learning.
\newblock In \emph{Arxiv 1802.01744}, 2018.
\newblock URL \url{https://arxiv.org/abs/1802.01744}.

\bibitem[Reitmann et~al.(2024)Reitmann, Mihaylova, Topp, and Kyrki]{reitmann2024shared}
S.~Reitmann, T.~Mihaylova, E.~A. Topp, and V.~Kyrki.
\newblock Conflict simulation for shared autonomy in autonomous driving.
\newblock In \emph{Companion of the 2024 ACM/IEEE International Conference on Human-Robot Interaction}, HRI '24, page 882–887, New York, NY, USA, 2024. Association for Computing Machinery.
\newblock ISBN 9798400703232.
\newblock \doi{10.1145/3610978.3640589}.
\newblock URL \url{https://doi.org/10.1145/3610978.3640589}.

\bibitem[Ropelato et~al.(2018)Ropelato, Z{\"u}nd, Magnenat, Menozzi, and van Dinther]{ropelato2018adaptive}
S.~Ropelato, F.~Z{\"u}nd, S.~Magnenat, M.~Menozzi, and Y.~van Dinther.
\newblock Adaptive tutoring on a virtual reality driving simulator.
\newblock \emph{ETH ZurichInternational SERIES on Information Systems and Management in Creative EMedia}, 2017\penalty0 (2):\penalty0 12--17, 2018.

\bibitem[Saadah and Saadah(2004)]{saadah2004autonomy}
M.~A. Saadah and L.~M. Saadah.
\newblock Autonomy and rehabilitation.
\newblock \emph{Neurosciences Journal}, 9\penalty0 (2):\penalty0 84--90, 2004.

\bibitem[Schrum et~al.(2022{\natexlab{a}})Schrum, Hedlund-Botti, and Gombolay]{schrum2022reciprocal}
M.~L. Schrum, E.~Hedlund-Botti, and M.~Gombolay.
\newblock Reciprocal {MIND} {MELD}: Improving learning from demonstration via personalized, reciprocal teaching.
\newblock In \emph{6th Annual Conference on Robot Learning}, 2022{\natexlab{a}}.
\newblock URL \url{https://openreview.net/forum?id=f_XmiyZcsjL}.

\bibitem[Schrum et~al.(2022{\natexlab{b}})Schrum, Hedlund-Botti, Moorman, and Gombolay]{schrum2022mind}
M.~L. Schrum, E.~Hedlund-Botti, N.~Moorman, and M.~C. Gombolay.
\newblock Mind meld: Personalized meta-learning for robot-centric imitation learning.
\newblock In \emph{2022 17th ACM/IEEE International Conference on Human-Robot Interaction (HRI)}, pages 157--165. IEEE, 2022{\natexlab{b}}.

\bibitem[Seita et~al.(2019)Seita, Chan, Rao, Tang, Zhao, and Canny]{seita2019zpd}
D.~Seita, D.~Chan, R.~Rao, C.~Tang, M.~Zhao, and J.~Canny.
\newblock Zpd teaching strategies for deep reinforcement learning from demonstrations.
\newblock \emph{arXiv preprint arXiv:1910.12154}, 2019.

\bibitem[Shiarlis et~al.(2018)Shiarlis, Wulfmeier, Salter, Whiteson, and Posner]{shiarlis2018taco}
K.~Shiarlis, M.~Wulfmeier, S.~Salter, S.~Whiteson, and I.~Posner.
\newblock {TACO}: Learning task decomposition via temporal alignment for control.
\newblock In J.~Dy and A.~Krause, editors, \emph{Proceedings of the 35th International Conference on Machine Learning}, volume~80 of \emph{Proceedings of Machine Learning Research}, pages 4654--4663. PMLR, 10--15 Jul 2018.
\newblock URL \url{https://proceedings.mlr.press/v80/shiarlis18a.html}.

\bibitem[Srivastava et~al.(2022)Srivastava, Biyik, Mirchandani, Goodman, and Sadigh]{srivastava2022motor}
M.~Srivastava, E.~Biyik, S.~Mirchandani, N.~Goodman, and D.~Sadigh.
\newblock Assistive teaching of motor control tasks to humans.
\newblock In S.~Koyejo, S.~Mohamed, A.~Agarwal, D.~Belgrave, K.~Cho, and A.~Oh, editors, \emph{Advances in Neural Information Processing Systems}, volume~35, pages 28517--28529. Curran Associates, Inc., 2022.

\bibitem[Sutton et~al.(1999)Sutton, Precup, and Singh]{sutton1999between}
R.~S. Sutton, D.~Precup, and S.~Singh.
\newblock Between mdps and semi-mdps: A framework for temporal abstraction in reinforcement learning.
\newblock \emph{Articial intelligence}, 112:\penalty0 181--211, 1999.

\bibitem[Tio and Varakantham(2023)]{tio2023training}
S.~Tio and P.~Varakantham.
\newblock Training reinforcement learning agents and humans with difficulty-conditioned generators.
\newblock In \emph{Second Agent Learning in Open-Endedness Workshop}, 2023.

\bibitem[Tzannetos et~al.(2024)Tzannetos, Kamalaruban, and Singla]{tzannetos2024proximal}
G.~Tzannetos, P.~Kamalaruban, and A.~Singla.
\newblock Proximal curriculum with task correlations for deep reinforcement learning.
\newblock \emph{arXiv preprint arXiv:2405.02481}, 2024.

\bibitem[Udupa et~al.(2023)Udupa, Kamat, and Menassa]{udupa2023shared}
S.~Udupa, V.~R. Kamat, and C.~C. Menassa.
\newblock Shared autonomy in assistive mobile robots: a review.
\newblock \emph{Disability and Rehabilitation: Assistive Technology}, 18\penalty0 (6):\penalty0 827--848, 2023.

\bibitem[Vainas et~al.(2019)Vainas, Bar-Ilan, Ben-David, Gilad-Bachrach, Lukin, Ronen, Shillo, and Sitton]{vainas2019gotsky}
O.~Vainas, O.~Bar-Ilan, Y.~Ben-David, R.~Gilad-Bachrach, G.~Lukin, M.~Ronen, R.~Shillo, and D.~Sitton.
\newblock E-gotsky: sequencing content using the zone of proximal development.
\newblock \emph{arXiv preprint arXiv:1904.12268}, 2019.

\bibitem[Vygotsky(1978)]{vygotsky1978mind}
L.~S. Vygotsky.
\newblock \emph{Mind in society: The development of higher psychological processes}, volume~86.
\newblock Harvard university press, 1978.

\bibitem[Wang et~al.(2022)Wang, Mu, Arumugam, Jaques, and Goodman]{wang2022zone}
R.~E. Wang, J.~Mu, D.~Arumugam, N.~Jaques, and N.~Goodman.
\newblock In the zone: Measuring difficulty and progression in curriculum generation.
\newblock In \emph{Deep Reinforcement Learning Workshop NeurIPS 2022}, 2022.

\bibitem[Wang et~al.(2020)Wang, Na, Cao, Gong, Xi, Xing, and Wang"]{wang2020review}
W.~Wang, X.~Na, D.~Cao, J.~Gong, J.~Xi, Y.~Xing, and F.-Y. Wang".
\newblock Decision-making in driver-automation shared control: A review and perspectives, 2020.
\newblock ISSN 2329-9266.

\bibitem[Williams et~al.(2024)Williams, Ali, Alcantara, Burghleh, Alghowinem, and Breazeal]{william2024doodlebot}
R.~Williams, S.~Ali, R.~Alcantara, T.~Burghleh, S.~Alghowinem, and C.~Breazeal.
\newblock Doodlebot: An educational robot for creativity and ai literacy.
\newblock In \emph{Proceedings of the 2024 ACM/IEEE International Conference on Human-Robot Interaction}, HRI '24, page 772–780, New York, NY, USA, 2024. Association for Computing Machinery.
\newblock ISBN 9798400703225.
\newblock \doi{10.1145/3610977.3634950}.
\newblock URL \url{https://doi.org/10.1145/3610977.3634950}.

\bibitem[Xing et~al.(2020)Xing, Huang, and Lv]{xing2020driver}
Y.~Xing, C.~Huang, and C.~Lv.
\newblock Driver-automation collaboration for automated vehicles: a review of human-centered shared control.
\newblock In \emph{2020 IEEE Intelligent Vehicles Symposium (IV)}, pages 1964--1971. IEEE, 2020.

\bibitem[Yu et~al.(2023)Yu, Xu, Li, and Hsu]{yu2023coach}
C.~Yu, Y.~Xu, L.~Li, and D.~Hsu.
\newblock Coach: Cooperative robot teaching.
\newblock In K.~Liu, D.~Kulic, and J.~Ichnowski, editors, \emph{Proceedings of The 6th Conference on Robot Learning}, volume 205 of \emph{Proceedings of Machine Learning Research}, pages 1092--1103. PMLR, 14--18 Dec 2023.
\newblock URL \url{https://proceedings.mlr.press/v205/yu23b.html}.

\bibitem[Zurek et~al.(2021)Zurek, Bobu, Brown, and Dragan]{zurek2021situational}
M.~Zurek, A.~Bobu, D.~S. Brown, and A.~D. Dragan.
\newblock Situational confidence assistance for lifelong shared autonomy.
\newblock In \emph{2021 IEEE International Conference on Robotics and Automation (ICRA)}, pages 2783--2789. IEEE, 2021.

\end{thebibliography}



\newpage
\newpage
~\newpage

\section{Vehicle Physics Control Parameters}

\begin{verbatim}
tire_front :
tire_friction=3.05, 
damping_rate=0.250000, 
max_steer_angle=69.999992, 
radius=37.000000,
max_brake_torque=3000, 
max_handbrake_torque=0.000000, 
lat_stiff_max_load=3.000000, 
lat_stiff_value=20.000000, 
long_stiff_value=3000.000000
        
tire_back : 
tire_friction=3.05, 
damping_rate=0.250000, 
max_steer_angle=0, 
radius=37.000000, 
max_brake_torque=3000, 
max_handbrake_torque=3000, 
lat_stiff_max_load=3.000000, 
lat_stiff_value=20.000000, 
long_stiff_value=3000.000000
        
Physics control for the car

torque_curve = 
- [0, 1000]
- [1890.760742, 900]
- [5729.577637, 700]
max_rpm = 8750
moi= 1.0
damping_rate_full_throttle= 0.150000
damping_rate_zero_throttle_clutch_engaged= 2.000000
damping_rate_zero_throttle_clutch_disengaged= 0.350000
use_gear_autobox= True
gear_switch_time= 0.000000
clutch_strength= 10.000000
final_ratio= 4.170000
mass= 1775.000000
drag_coefficient=0.300000
center_of_mass: [0.500000, 0.000000, -0.300000]   
steering_curve = 
- [0, 1.0]
- [20, 0.9]
- [60, 0.8]
- [120, 0.7]
\end{verbatim}

\section{User Study Instructions}

{\includegraphics[width=0.45\textwidth]{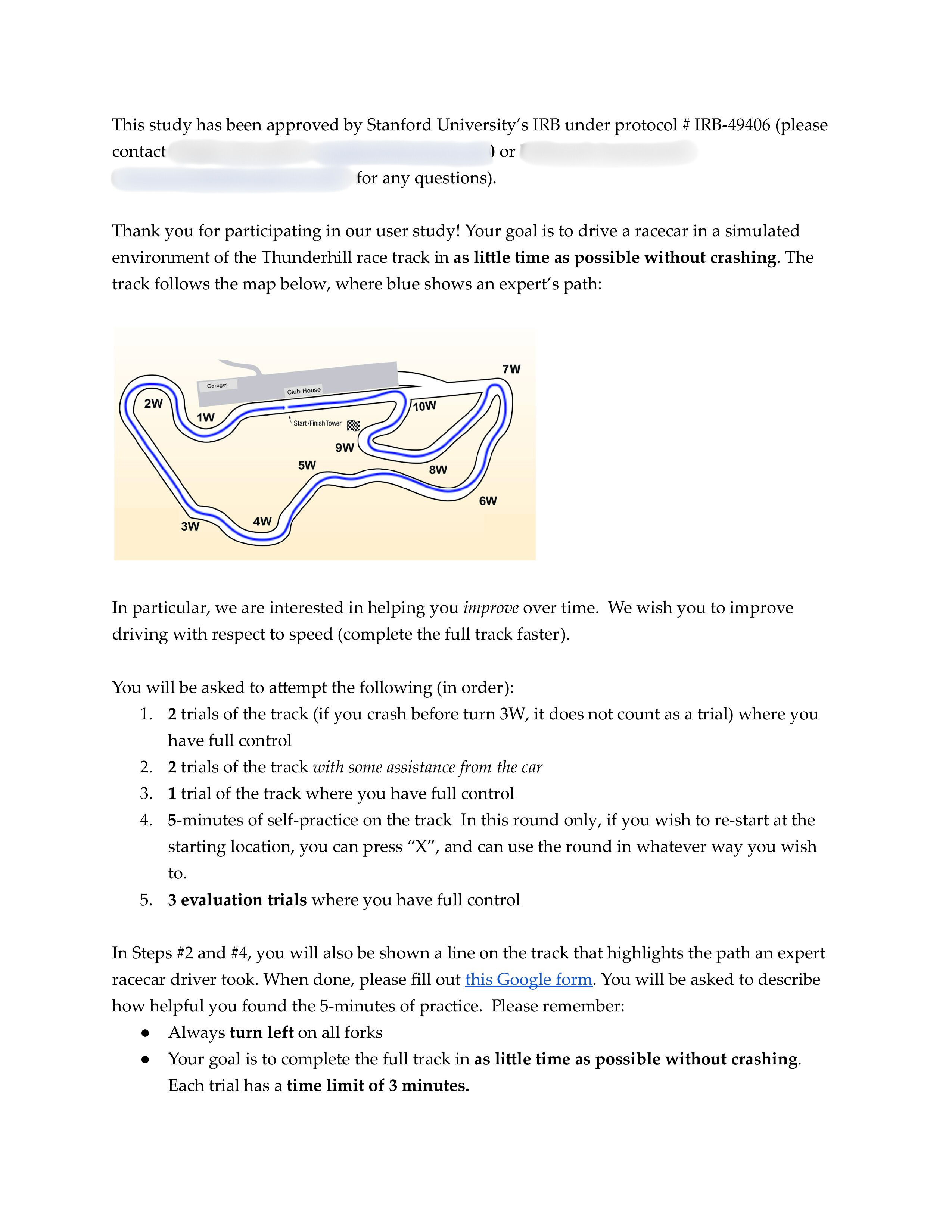}}

\section{Post-Study Google Form}

The purpose of this Google form was to gather participant feedback on their experiences with CARLA after the nine trials and self-practice sessions. The form collects data on the participants' driving habits, video game usage, the effectiveness of assistance provided during the trials, and their overall learning process. All questions were mandatory to answer.
{\includegraphics[width=0.45\textwidth]{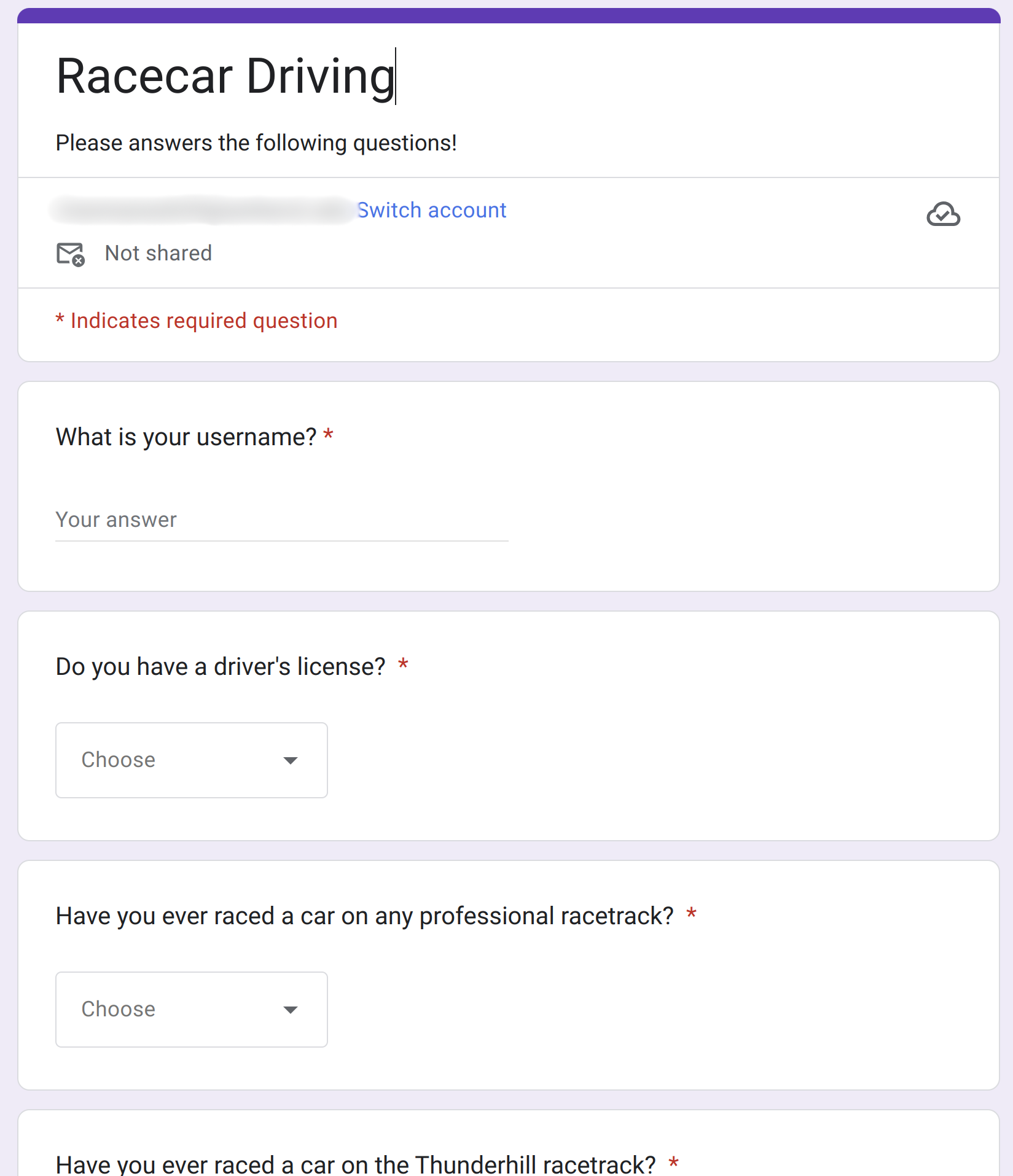}}

\section{Google Form: Racecar Driving Questionnaire}

Please answer the following questions!

\begin{enumerate}
    \item \textbf{What is your username?} \\
    Your answer

    \item \textbf{Do you have a driver's license?} \\
    (Choose Yes/No)

    \item \textbf{Have you ever raced a car on any professional racetrack?} \\
    (Choose Yes/No)

    \item \textbf{Have you ever raced a car on the Thunderhill racetrack?} \\
    (Choose Yes/No)

    \item \textbf{Around how many hours do you spend driving per week?} \\
    Your answer

    \item \textbf{Around how many hours do you spend playing video games per week?} \\
    Your answer

    \item \textbf{How would you describe the assistance you received in Trials 3 and 4 (before the practice session)?} \\
    Your answer

    \item \textbf{How helpful did you find the assistance you received in Trials 3 and 4 (before the practice session)?} \\
    Not helpful at all [1] [2] [3] [4] [5] Very helpful

    \item \textbf{To what degree did the assistance you received in Trials 3 and 4 (before the practice session) challenge you?} \\
    Did not challenge me at all [1] [2] [3] [4] [5] Challenged me a lot

    \item \textbf{How helpful did you find the 5-minute practice session?} \\
    Not helpful at all [1] [2] [3] [4] [5] Very helpful

    \item \textbf{To what degree did the assistance you received in the 5-minute practice session challenge you?} \\
    Did not challenge me at all [1] [2] [3] [4] [5] Challenged me a lot

    \item \textbf{How did you use the 5-minute practice session? (e.g. did you focus on improving any particular part of the task?)} \\
    Your answer

    \item \textbf{What is something you learned about the task over time that helped you?} \\
    Your answer

    \item \textbf{What other feedback do you have about the study?} \\
    Your answer
\end{enumerate}

\section{Feedback Dataset}

The dataset used for the skill discovery component was created from a subset of the data collected looking at naturalistic teaching.  In this data set, participants (n=15) received up to 90 minutes of performance-driving coaching on a driving simulator from a professional coach. The coach was instructed to 1) help students become better drivers and 2) use primarily verbal feedback. Participants drove around a virtual version of the  Thunderhill West track, located in Northern California. Each 15 minutes students filled out surveys to assess their cognitive load and emotional state. Each 15 minutes, the coach filled out surveys assessing the student's skill and improvement.  We used data from a single participant to use for model validation. 

This study was reviewed and approved by an IRB (name to be entered after review). Prior to participating, participants provided with written informed consent. Afterwards, the coach provided each participant with a simulator safety and conduct lesson and drove a site lap. The participants were then prompted to drive two laps around to gauge their baseline driving skill before the training began. Then, participants, depending on time, completed three to five 15-minute sessions with the coach, where the lessons were guided by what skills the coach, and occasionally the students, thought the student needed to work on. Each session was divided by both the participant and the coach filling out a survey about how the session went. 

After the sessions were completed, the participant and coach filled out a final survey about the overall experience of the training. The study ended with the participant completing 2 final laps around the track with no coaching to gauge how much their skill improved. 

With this study structure, we were able to collect the trajectory and behavioral data for each participant, which was crucial for helping validate the model.

\vspace{12pt}

\end{document}